\documentclass[10pt,twocolumn,letterpaper]{article}

\usepackage[pagenumbers]{cvpr} % To force page numbers, e.g. for an arXiv version

\usepackage{graphicx}
\usepackage{amsmath}
\usepackage{amssymb}
\usepackage{booktabs}
\usepackage{multirow}
\usepackage{float}
\usepackage{tabularx} % for 'tabularx' env. and 'X' col. type
\usepackage{ragged2e} % for \RaggedRight macro
\usepackage{booktabs} % for \toprule, \midrule etc macros
\usepackage[accsupp]{axessibility}
\usepackage[pagebackref,breaklinks,colorlinks]{hyperref}

\usepackage[capitalize]{cleveref}
\crefname{section}{Sec.}{Secs.}
\Crefname{section}{Section}{Sections}
\Crefname{table}{Table}{Tables}
\crefname{table}{Tab.}{Tabs.}

\def\cvprPaperID{1791} % *** Enter the CVPR Paper ID here
\def\confName{CVPR}
\def\confYear{2023}

\begin{document}

\title{OpenGait: Revisiting Gait Recognition Toward Better Practicality}

% \and
\author{
Chao Fan$^{1,2}$, 
Junhao Liang$^{1,2}$, 
Chuanfu Shen$^{3,1}$, 
Saihui Hou$^{4,5}$, \\
Yongzhen Huang$^{4,5}$, 
Shiqi Yu$^{1,2}$\thanks{Corresponding Author} \\
{\normalsize $^1$ Department of Computer Science and Engineering, Southern University of Science and Technology}\\
{\normalsize $^2$ Research Institute of Trustworthy Autonomous System, Southern University of Science and Technology} \\
{\normalsize $^3$ Department of Industrial and Manufacturing Systems Engineering, The University of Hong Kong} \\
{\normalsize $^4$ School of Artificial Intelligence, Beijing Normal University} 
{\normalsize $^5$ WATRIX.AI} \\
{\tt\scriptsize \{12131100, 12132342, 11950016\}@mail.sustech.edu.cn, \{housaihui, huangyongzhen\}@bnu.edu.cn, yusq@sustech.edu.cn}
}

\maketitle
% \end{}
%%%%%%%%% ABSTRACT
\begin{abstract}
Gait recognition is one of the most critical long-distance identification technologies and increasingly gains popularity in both research and industry communities. 
Despite the significant progress made in indoor datasets, much evidence shows that gait recognition techniques perform poorly in the wild.
More importantly, we also find that some conclusions drawn from indoor datasets cannot be generalized to real applications.
Therefore, the primary goal of this paper is to present a comprehensive benchmark study for better practicality rather than only a particular model for better performance.
To this end, we first develop a flexible and efficient gait recognition codebase named OpenGait. 
Based on OpenGait, we deeply revisit the recent development of gait recognition by re-conducting the ablative experiments.
Encouragingly,we detect some unperfect parts of certain prior woks, as well as new insights.
Inspired by these discoveries, 
we develop a structurally simple, empirically powerful, and practically robust baseline model, GaitBase.
Experimentally, we comprehensively compare GaitBase with many current gait recognition methods on multiple public datasets, and the results reflect that GaitBase achieves significantly strong performance in most cases regardless of indoor or outdoor situations.
Code is available at \url{https://github.com/ShiqiYu/OpenGait}.
\end{abstract}

%%%%%%%%% BODY TEXT
\vspace{-1.5em}
\section{Introduction}
\label{sec:intro}
\begin{figure}[tb]
\centering
\includegraphics[height=6.5cm]{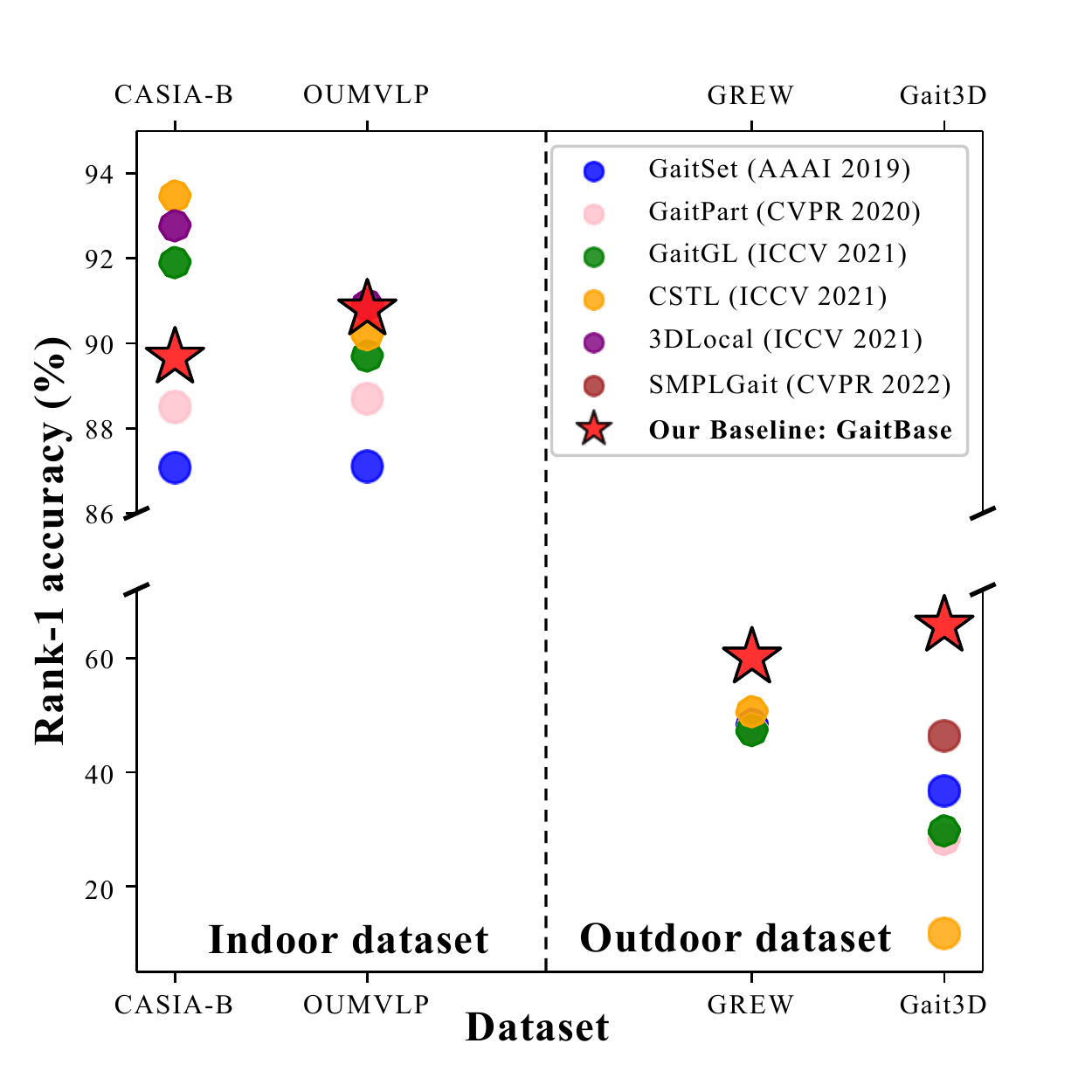}
\caption{Performance of popular models and ours baseline on 4 major gait datasets~\cite{chao2019gaitset,fan2020gaitpart,lin2021gait,huang2021context,huang20213d,zheng2022gait3d}. The left two are indoor datasets~\cite{yu2006framework,takemura2018multi}, the right two are outdoor datasets~\cite{zhu2021gait,zheng2022gait3d}.
}
\vspace{-1.5em}
\label{fig:perf}
\end{figure}
Gait recognition  has recently gained growing interest from the vision research community.
It utilizes the physiological and behavioral characteristics from walking videos to authenticate individuals's identities~\cite{wang2003silhouette}.
Compared with other biometrics, 
\textit{e.g.}, face, fingerprint, and iris, 
gait patterns can be captured from a distance in uncontrolled settings, without requiring any physical contact.
As a walking behavior, 
gait is also hard to disguise and thus promisingly robust for usual subject-related covariates, 
such as dressing, carrying, and standing conditions.
These advantages make gait recognition suitable for public security applications, 
\textit{e.g.}, criminal investigation, and suspect tracking~\cite{wu2016comprehensive}.

With the boom of deep learning, gait recognition in the laboratory~\cite{yu2006framework,takemura2018multi} has achieved significant progress ~\cite{chao2019gaitset,fan2020gaitpart,lin2021gait} over the last decade.
However, much evidence~\cite{zhu2021gait, zheng2022gait3d} reveal that gait recognition techniques may not perform optimally in the wild.
As shown in Figure~\ref{fig:perf}, 
most existing gait recognition methods suffer an over 40$\%$ accuracy degradation when transitioning from indoor to outdoor datasets. 
Typically, this performance gap should be mainly caused by real-world noisy factors, such as complex occlusion, background variation, and illumination changes.

Nevertheless, our further ablative study shows that this situation is not unique, as many of the conclusions drawn in prior works vary with different datasets. 
Therefore, besides proposing an improved model for better performance, the primary objective of this paper is to present a comprehensive benchmark study to revisit gait recognition for enhanced practicality. To this end, we make efforts in the following three aspects.

Firstly, to the best of our knowledge, previous works mainly develop the models on their code repository and rely heavily on the indoor gait datasets, particularly CASIA-B~\cite{yu2006framework} and OU-MVLP~\cite{takemura2018multi}.
To accelerate the real-world applications, we appeal to pay more attention to outdoor gait datasets, such as GREW~\cite{zhu2021gait} and Gait3D~\cite{zheng2022gait3d}.
Additionally, this paper also considers building a unified evaluation platform, which covers the various state-of-the-art methods and testing datasets, is highly desired.
Accordingly, we propose a flexible and efficient gait recognition codebase with PyTorch~\cite{paszke2017automatic} and name it \textbf{OpenGait}.

To ensure extensibility and reusability, 
OpenGait supports  the following features:
(1)~\textbf{Various datasets}, 
\textit{e.g.}, 
the indoor CASIA-B~\cite{yu2006framework} and OU-MVLP~\cite{takemura2018multi}, the outdoor GREW~\cite{zhu2021gait} and Gait3D~\cite{zheng2022gait3d}.
(2)~\textbf{State-of-the-art methods}, 
\textit{e.g.}, 
GaitSet~\cite{chao2019gaitset}, GaitPart~\cite{fan2020gaitpart}, GLN~\cite{hou2020gait}, GaitGL~\cite{lin2021gait},
% CSTL~\cite{huang2021context},
SMPLGait~\cite{zheng2022gait3d}, GaitEdge\cite{liang2022gaitedge}, and so on.
% future ones.
(3)~\textbf{Multiple popular frameworks},  
\textit{e.g.}, 
the end-to-end, multi-modality, and contrastive learning paradigms.
Recently, OpenGait has been widely employed in two of the major international gait recognition competitions, \textit{i.e.}. HID~\cite{yu2022hid}\footnote{
HID 2023: \url{https://hid2023.iapr-tc4.org}}, and GREW~\cite{zhu2021gait}.
Encouragingly, all of the top-10 winning teams at HID2022~\cite{yu2022hid} have utilized OpenGait as their codebase and extended OpenGait to develop new solutions.

Based on OpenGait, 
we reproduce many progressive methods~\cite{chao2019gaitset,fan2020gaitpart,lin2021gait}, and the results have been shown in Figure~\ref{fig:perf}.
More importantly, we conduct a comprehensive re-evaluation of various commonly accepted conclusions by re-implementing the ablation study on recently-built outdoor gait datasets. 
To our surprise, we find that the MGP branch proposed by GaitSet~\cite{chao2019gaitset}, 
the FConv proposed by GaitPart~\cite{fan2020gaitpart},
the local feature extraction branch proposed by GaitGL~\cite{lin2021gait}, 
and the SMPL branch proposed by SMPLGait~\cite{zheng2022gait3d},
do not exhibit superiority on the outdoor datasets. 
Moreover, our thorough exploration of potential causes reveals several hidden limitations of prior gait research, such as the lack of comprehensive ablation study, outdoor dataset evaluation, and a strong backbone network.

Inspired by the above discoveries, we develop a simple yet strong baseline model, named \textbf{GaitBase}.
Specifically, GaitBase is composed of several essential parts, some of which are simple and commonly utilized rather than intricately constructed modules.
Even no bells or whistles, 
GaitBase can achieve comparable or even superior performance on indoor gait datasets.
As to the datasets in the wild, 
GaitBase outperforms recently proposed methods and reaches a new state-of-the-art.
Furthermore, we also conduct a comprehensive study to verify that GaitBase is a structurally simple, experimentally powerful, and empirically robust baseline model for gait recognition.

In summary, this paper contributes future works from three aspects: 
(1) OpenGait, a unified and extensible platform, is built to facilitate the comprehensive study of gait recognition.
(2) We deeply revisit the recent gait recognition developments and consequently bring many new insights for further gait recognition research.
(3) We provide a structurally simple and experimentally powerful baseline model, GaitBase, which can inspire the future design of gait recognition algorithms.
We hope the works in the paper can inspire researchers to develop more advanced gait recognition methods for real-world applications.

\label{sec:opengait}
\begin{figure*}[tb]
\centering
\includegraphics[height=3.5cm]{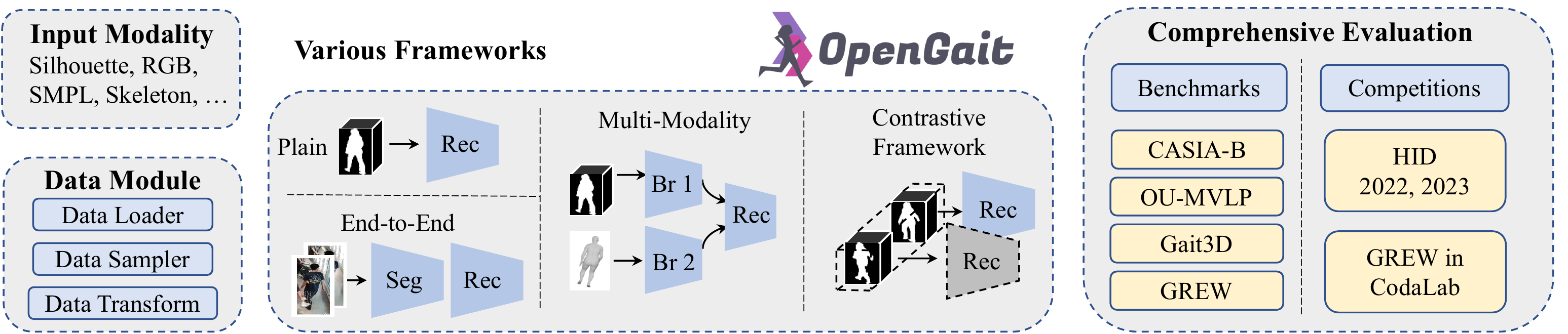}
\caption{The design principles of proposed codebase \textbf{OpenGait}.
Seg is for \textit{segmentation}, 
Rec is for \textit{recognition}, 
and Br is for \textit{branch}.
}
\label{fig:framework}
\vspace{-1.5em}
\end{figure*}

\vspace{-0.5em}
\section{Related Work}
\vspace{-0.5em}
According to the classical taxonomy, 
gait recognition methods can be roughly grouped to two categories, 
including the model-based and appearance-based methods.

\noindent
\textbf{Model-based Gait Recognition} methods~\cite{liao2020model,teepe2021gaitgraph,li2020end} tend to take the estimated underlying structure of human body as input, such as 2D/3D pose and SMPL~\cite{loper2015smpl} model.
Specifically, PoseGait~\cite{liao2020model} used 3D body pose and human prior knowledge to overcome the changes in clothing, GaitGraph~\cite{teepe2021gaitgraph} introduced graph convolutional network for 2D skeleton-based gait representation learning, and HMRGait~\cite{li2020end} fine-tuned a pre-trained human mesh recovery network to construct the end-to-end SMPL-based model.
It is also worth mentioning some model-based multi-modality frameworks, such as SMPLGait~\cite{zheng2022gait3d} that exploited the 3D geometrical information from SMPL model to enhance the gait appearance feature learning, and BiFusion~\cite{peng2021learning} that integrated skeletons and silhouettes to capture the rich gait spatio-temporal features.
Our OpenGait supports all the aforementioned human body models. 
However, while these human models are theoretically robust against noisy factors like carrying and dressing, they often struggle with low-resolution situations and, as a result, may lack practicality in some real-world scenarios.

\noindent
\textbf{Appearance-based Gait Recognition} methods directly learn shape features from the input video, 
which suit the low-resolution conditions and thus attract increasing attention.
With the boom of deep learning, 
most current appearance-based works focus on spatial feature extraction and gait temporal modeling.
Specifically, 
GaitSet~\cite{chao2019gaitset} innovatively regarded the gait sequence as a set and utilized a maximum function to compress the sequence of frame-level spatial features. 
Thanks to its simplicity and effectiveness, 
GaitSet has became one of the most influential gait recognition works in recent years.
GaitPart~\cite{fan2020gaitpart} carefully explored the local details of input silhouette and modeled the temporal dependencies by Micro-motion Capture Module~\cite{fan2020gaitpart}.
GaitGL~\cite{lin2021gait} argued that the spatially global gait representations often neglect the details, and the local region-based descriptors cannot capture the relations among neighboring parts, thus developing the global and local convolution layer~\cite{lin2021gait}.
CSTL~\cite{huang2021context} focused on temporal features in three scales to obtain motion representation according to the temporal contextual information.
3DLocal~\cite{huang20213d} wanted to extract the limb features through 3D local operations at adaptive scales.

In addition to the above, 
many other outstanding works inspire us a lot as well, 
\textit{e.g.}, 
GaitEdge~\cite{liang2022gaitedge} designed an edge-trainable intermediate modality to build the end-to-end gait recognition framework, 
and GaitSSB~\cite{fan2022learning} collected millions of unlabelled gait sequences and learned the general gait representation with a contrastive framework.

This paper introduces the OpenGait codebase, which is compatible with almost all of the aforementioned methods. Additionally, as research directions shift from indoor to outdoor environments, we re-analyzed many of these methods from an experimental perspective and gained new insights. Finally, we developed a simple yet powerful baseline model, GaitBase, which outperforms prior works by a significant margin on outdoor datasets.

\noindent
\textbf{Gait Datasets} are also crucial for gait recognition research, and among them, CASIA-B~\cite{yu2006framework} and OU-MVLP~\cite{takemura2018multi} are two of the most widely-used indoor datasets.
They were proposed in 2006 and 2018, respectively, and captured by requiring subjects to walk around the laboratory, which significantly differs from the real-world scenarios.
In response to the need for real-world applications, GREW~\cite{zhu2021gait} and Gait3D~\cite{zheng2022gait3d} were collected in the wild in 2021 and 2022, respectively.
However, most existing works only verify their effectiveness on indoor datasets, which poses a high risk of vulnerability in practical usage.

\noindent
\textbf{Revisit Deep Gait Recognition.}
Recently, several survey works~\cite{filipi2022gait,sepas2022deep,shen2022comprehensive} have investigated almost all published papers on deep gait recognition. However, a more comprehensive and detailed analysis of these methods is still lacking. In some other fields, there are critical papers revisiting previous methods, such as recommendation systems~\cite{ferrari2019we}, metric learning~\cite{musgrave2020metric}, and unsupervised domain adaptation~\cite{musgrave2021unsupervised}. OpenGait aims to address this gap in the field of gait recognition by experimentally reviewing recent works and highlighting some concerns about their robustness.

\noindent
\textbf{Codebase.}
There are many works providing infrastructure in the deep learning research community, 
such as a codebase for the specific research topic. 
For example, Amos \textit{et al.}~\cite{amos2016openface} proposed OpenFace, a face recognition library that bridges the gap between public face recognition systems and industry-leading private systems. 
In the field of object detection, 
a PyTorch toolbox called MMDetection~\cite{mmdetection} supports almost all popular detection methods, 
providing a convenient platform for systematic comparison.
With the rapid development of gait recognition, 
the need for an infrastructure code platform has become increasingly prominent. 

\vspace{-0.5em}
\section{OpenGait}
\vspace{-0.5em}
Over the past few years, 
numerous new frameworks and evaluation datasets have emerged for gait recognition. 
However, the lack of a unified and fair evaluation platform cannot be overlooked.
To facilitate academic research and practical applications, 
this paper presents a PyTorch-based~\cite{paszke2019pytorch} toolbox, \textbf{OpenGait}, as a reasonable and dependable solution to address this issue.

\vspace{-0.5em}
\subsection{Design Principles of OpenGait}
\vspace{-0.5em}
As shown in Figure~\ref{fig:framework}, our developed OpenGait covers the following highlight features.

\noindent
\textbf{Compatibility with Diverse Gait Modalities.} 
Gait recognition has a variety of input modalities. 
The usual ones include silhouette images, 2D/3D skeletons, and more recently emerging modalities include SMPL parameters~\cite{li2020end,zheng2022gait3d} and RGB images~\cite{song2019gaitnet,liang2022gaitedge}. 
Existing open-source repositories mostly only support one of these modalities, 
while our OpenGait is designed to support all of these modalities.

\noindent
\textbf{Compatibility with Various Frameworks.} 
Currently, more and more novel gait recognition methods have emerged, 
such as multi-modalities~\cite{zheng2022gait3d}, end-to-end~\cite{song2019gaitnet,li2020end,liang2022gaitedge}, and contrastive learning~\cite{fan2022learning}. 
As mentioned above, most open-source methods narrowly focus on their own models, so extending to multiple frameworks is difficult.
Fortunately, OpenGait supports all of the above frameworks.

\noindent
\textbf{Support for Various Evaluation Datasets.} 
OpenGait is a comprehensive toolbox that includes datasets commonly used by researchers. 
We offer full support for indoor gait datasets such as CASIA-B and OU-MVLP, as well as newly proposed outdoor wild datasets GREW and Gait3D. 
OpenGait provides a range of uniquely designed functions for these datasets, from data preprocessing to final evaluation. 
In addition, it is noteworthy that two of the major international competitions, HID~\cite{yu2022hid} and GREW~\cite{zhu2021gait}, are compatible with OpenGait.
Many winning teams have utilized OpenGait to develop new solutions.

\noindent
\textbf{Support for State-of-the-arts.} 
We reproduce many previous state-of-the-art methods, including GaitSet~\cite{chao2019gaitset}, GaitPart~\cite{fan2020gaitpart}, GLN~\cite{hou2020gait}, GaitGL~\cite{lin2021gait},Gait3D~\cite{zheng2022gait3d}, SMPLGait\cite{zheng2022gait3d}, GaitEdge~\cite{liang2022gaitedge} and GaitSSB~\cite{fan2022learning}. 
Reproduced performances are comparable to or even better than the results reported by the original papers. 
Rich official examples can help beginners get started with the project better and faster. 
In addition, this also provides an infrastructure for a more comprehensive and systematic comparison later. 

\vspace{-0.5em}
\subsection{Main Modules}
\vspace{-0.5em}
\label{sec:modules}
Technically, we follow the design of most PyTorch deep learning projects and divide OpenGait into three modules, \textit{data}, \textit{modeling}, and \textit{evaluation}, as shown in Figure~\ref{fig:framework}. 
% Their details are separately introduced below.

\label{sec:doub}

\textit{Data module}
contains data loader, data sampler and data transform, which are responsible for loading, sampling, and pre-processing the input data flow respectively.

\textit{Modeling module}
is built on top of a base class (\texttt{BaseModel}) that pre-defines many behaviors of the deep model during training and testing phases, including optimization and inference.
The four essential components of current gait recognition algorithms, namely \textit{backbone}, \textit{neck}, \textit{head}, and \textit{loss}, can be customized in this class.

\textit{Evaluation module}
is used to evaluate the obtained model. 
It is well known that different datasets have various evaluation protocols, and we unify them into OpenGait to free researchers from these tedious details. 

\vspace{-0.5em}
\section{Revisit Deep Gait Recognition}
\vspace{-0.5em}
\label{sec:revisit}
With the help of OpenGait, 
we can revisit several typical gait recognition methods comprehensively.
Some insights different from those presented in the original papers have been found from our fair ablation studies.

\vspace{-0.5em}
\subsection{Experimental Recheck on Previous Methods}
\vspace{-0.5em}
We notice that most previous works only verify the effectiveness on the indoor gait datasets, \textit{i.e.}, CASIA-B\cite{yu2006framework} and OU-MVLP\cite{takemura2018multi}, 
and the further ablation study is only conducted based on the CASIA-B.
In this section, 
excluding reproducing the officially stated performance, 
we re-conduct some ablation studies based on the newly-built outdoor dataset, \textit{i.e.}, Gait3D, 
to experimentally check the algorithm's robustness for practical gait data.

\begin{table}[t]
\centering
\caption{The effect of MGP and Multi-scale HPP in GaitSet~\cite{chao2019gaitset}. In the original GaitSet~\cite{chao2019gaitset}, all the obtained partial vectors will be concatenated into just one feature vector used for the final evaluation. However, following works~\cite{fan2020gaitpart, lin2021gait, zheng2022gait3d} find this concatenation unnecessary and take the average of partial distance for evaluation. Therefore, OpenGait follows this manner and reproduces a better performance than the official results.}
\vspace{-0.5em}
\resizebox{1.0\columnwidth}{!}{
\begin{tabular}{c|c|ccc|cc} 
% \hline
\toprule
\multirow{2}{*}{MGP}      & \multirow{2}{*}{Multi-scale HPP} & \multicolumn{3}{c|}{CASIA-B}                  & \multicolumn{2}{c}{Gait3D}                       \\ 
\cline{3-7}
                          &                              & NM            & BG            & CL            & R-1                    & R-5                     \\ 
\hline\hline
\multirow{2}{*}{\checkmark}         & $\times$                           & \textbf{95.9} & 90.3          & \textbf{74.2} & 44.3                   & 64.7                    \\ 
\cline{2-7}
                          & \checkmark                           & 95.8          & 90.4          & 73.2          & 44.3                   & 64.4                    \\ 
\hline
\multirow{2}{*}{$\times$} & $\times$                           & 95.3          & \textbf{90.5} & 74.0          & \textbf{\textbf{45.8}} & \textbf{\textbf{65.1}}  \\ 
\cline{2-7}
                          & \checkmark                           & 94.5          & 89.1          & 72.3          & 43.7                   & 63.8                    \\
% \hline
\bottomrule
\end{tabular}
}
\label{tab:mgp}
\vspace{-0.5em}
\end{table}

\begin{figure*}[tb]
\centering
\includegraphics[height=6.5cm]{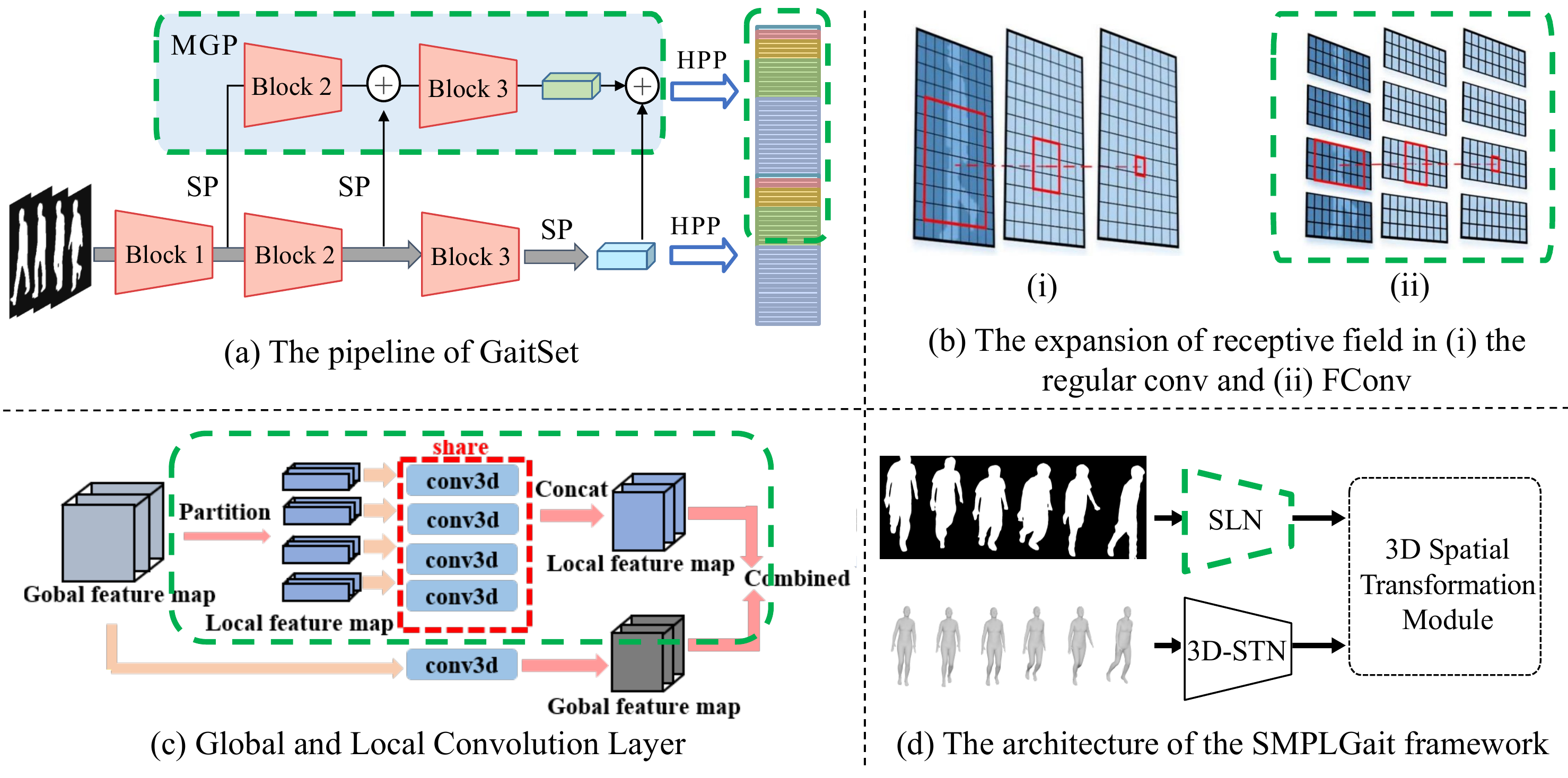}
\caption{The key modules of four previous SOTA methods. 
The modules enclosed by dotted green lines are the ones we  remove or replace for comparison. 
(a): \textbf{MGP} and multi-scale mechanism in \textbf{HPP} are removed~\cite{chao2019gaitset}.
(b): FConvs are replaced with the regular convolution layers~\cite{fan2020gaitpart}. 
(c): The upper local feature branch is replaced with an independent conv3d that is identical to the lower global branch~\cite{lin2021gait}. 
(d): \textbf{SLN} branch is replaced with our stronger ResNet-like backbone~\cite{zheng2022gait3d}.}
\label{fig:flaws}
\vspace{-1.5em}
\end{figure*}

\noindent \textbf{Re-conduct  Ablation Study on GaitSet.}
With taking the silhouettes as input, 
GaitSet~\cite{chao2019gaitset} treats the gait sequence as an unordered set and uses a simple maximum pooling function along the temporal dimension, called Set Pooling (SP), to generate the set-level understanding of the entire input gait sequence.
GaitSet~\cite{chao2019gaitset} provides insights for many subsequent works thanks to its simplicity and effectiveness,   
However, 
we find that the other two important components in GaitSet~\cite{chao2019gaitset}, 
namely the parallel Multi-layer Global Pipeline (MGP) and pyramid-like Horizontal Pyramid Pooling (HPP)~\cite{fu2019horizontal},
do not work well enough on both the indoor CASIA-B and outdoor Gait3D datasets. 
Specifically, as shown in Figure~\ref{fig:flaws} (a), 
MGP can be regarded as an additional branch that aggregates the hierarchical set-level characteristics. HPP~\cite{fu2019horizontal} follows the fashion feature pyramid structure aiming to extract multi-scale part-level representations. 
As shown in Table~\ref{tab:mgp}, 
if we strip MGP or remove the multi-scale mechanism in HPP from the official GaitSet~\cite{chao2019gaitset},  
the obtained model could reach the same or even better performance on both CASIA-B and Gait3D, saving over 80\% training weights.
This result indicates that the set-level characteristics extracted by bottom convolution blocks may be hard to benefit the final gait representation.
Besides, 
the multi-scale mechanism in HPP also provides no extra discriminative features. 
The cause may be that the employed statistical pooling functions are too weak to learn additional knowledge from various-scale human body parts.

\begin{table}[htbp]
\centering
\caption{The effect of FConv in GaitPart~\cite{fan2020gaitpart}. }
\vspace{-0.5em}
\resizebox{.70\columnwidth}{!}{
\begin{tabular}{c|ccc|cc} 
% \hline
\toprule
\multirow{2}{*}{FConv} & \multicolumn{3}{c|}{CASIA-B}         & \multicolumn{2}{c}{Gait3D}  \\ 
\cline{2-6}
                        & NM            & BG            & CL            & R-1           & R-5                  \\ 
\hline\hline
\checkmark                & \textbf{96.2} & \textbf{91.5} & \textbf{78.7} & 29.2          & 48.6                 \\ 
\hline
$\times$               & 95.6          & 88.4          & 76.1          & \textbf{36.2} & \textbf{57.0}        \\
% \hline
\bottomrule
\end{tabular}
}
% The experiments are conducted on CASIA-B and Gait3D}
\label{tab:fconv}
\vspace{-0.5em}
\end{table}

\noindent \textbf{Re-conduct  Ablation Study on GaitPart.} 
One of the core contributions of GaitPart~\cite{fan2020gaitpart} is to point out the importance of local details with the proposed Focal Convolution Layer (FConv).  
Figure~\ref{fig:flaws} (b) shows the receptive field's expansion of the top-layer neuron in the network composed of the regular and focal convolution layers.
Technically, 
FConv splits the input feature map into several parts horizontally and then performs a regular convolution over each part separately. 
As shown in Table \ref{tab:fconv}, we get a much higher performance (Rank-1: +7.0\%) on Gait3D with changing FConv to the regular convolution layer.
This phenomenon exhibits that the extraction of gait features may be seriously affected by directly splitting the feature map due to the low-quality segmentation of wild data.
The recently popular shifted window mechanism~\cite{liu2021swin} may address this issues.
Besides, 
since GaitPart requires sequential frames as input, 
some usual real-world factors in the temporal dimension may negatively impact the final performance as well,  
such as the inevitable frame drop and walking speed changes.

\begin{table}[t]
\centering
\caption{The effect of local branch in GaitGL~\cite{lin2021gait}.}
\vspace{-0.5em}
\resizebox{.80\columnwidth}{!}{
\begin{tabular}{c|ccc|cc} 
% \hline
\toprule
\multirow{2}{*}{Local Branch} & \multicolumn{3}{c|}{CASIA-B}         & \multicolumn{2}{c}{Gait3D}  \\ 
\cline{2-6}
                        & NM            & BG            & CL            & R-1           & R-5                  \\ 
\hline\hline
\checkmark       & \textbf{97.4} & \textbf{94.5} & \textbf{83.6} & 31.4          & 50.0                 \\ 
\hline
$\times$      & 97.1          & 93.7          & 81.9          & \textbf{32.2} & \textbf{52.5}        \\
% \hline
\bottomrule
\end{tabular}
}
% GL for Global and Local}
\label{tab:glfe}
\vspace{-1.0em}
\end{table}

\noindent \textbf{Re-conduct  Ablation Study on GaitGL.}
GaitGL~\cite{lin2021gait} argues that the spatially global-level gait representations often neglect the details of input gait frames.
At the same time, the local region-based descriptors cannot capture the relations among neighboring parts, thus developing the global and local convolution layer.
As shown in Fig~\ref{fig:flaws} (c), 
the local branch can be regarded as the FConv~\cite{fan2020gaitpart} employing 3D convolution, 
while the global branch is a standard 3D convolution layer.
Similar to GaitPart~\cite{fan2020gaitpart}, 
as shown in Table \ref{tab:glfe},
removing the local branch can achieve better performance on the outdoor Gait3D dataset.

\noindent \textbf{Re-conduct  Ablation Study on SMPLGait.}
As shown in Figure~\ref{fig:flaws} (d), 
SMPLGait~\cite{zheng2022gait3d} consists of two elaborately-designed branches,
\textit{i.e.}, 
silhouette (SLN) and SMPL (3D-STN) branches.
They are respectively used for 2D appearance extraction and 3D knowledge learning. 
SMPLGait~\cite{zheng2022gait3d} takes advantage of the 3D mesh data available in Gait3D~\cite{zheng2022gait3d} and achieves a gain effect on the top of the silhouette branch.  
However, 
as shown in Table~ \ref{tab:smpl}, 
our experiment demonstrates that the proposed SMPL branch not provides apparent promotion when we give the silhouette branch a strong backbone network.
\textit{e.g.}, 
ResNet-like network that will be built into our strong baseline.

In our view, there are three possible reasons causing the failure of the SMPL branch: a). Though the SMPL model is usually visualized as a dense mesh, its feature vector only possesses tens of dimensions that present the relatively sparse characterization of body shape and posture, making it challenging to enhance the fine-grained description of gait patterns. b). Since the SMPL model is not recognition-oriented, purposefully fine-tuning it may be more optimal than directly utilizing it to depict the subtle individual characteristics~\cite{li2020end}. c). In the wild, estimating an accurate SMPL model that finely captures body shape and posture from a single RGB camera is still challenging. In a nutshell, introducing 3D geometrical information from the SMPL model to benefit gait representation learning is well worth further exploration.

\begin{table}[t]
\centering
\caption{The effect of SMPL branch in SMPLGait~\cite{zheng2022gait3d}. 
SLN presents silhouette branch used in Gait3D. 
ResNet-like network is the backbone of our GaitBase.}
\vspace{-0.5em}
\resizebox{.82\columnwidth}{!}{
\begin{tabular}{c|cc|cc}
% \hline
\toprule
\multirow{3}{*}{SMPL branch} & \multicolumn{4}{c}{Silhouette branch}                                     \\ 
\cline{2-5}
                        & \multicolumn{2}{c|}{SLN} & \multicolumn{2}{c}{ResNet-like Network}  \\ 
\cline{2-5}
                        & R-1           & R-5               & R-1           & R-5                   \\ 
\hline\hline
\checkmark                & \textbf{46.3} & \textbf{64.5}     & 55.2          & \textbf{75.7}         \\ 
\hline
$\times$  & 42.9         & 63.9             & \textbf{56.5} & 75.2                  \\
% \hline
\bottomrule
\end{tabular}
}
\label{tab:smpl}
\vspace{-1.0em}
\end{table}

\vspace{-0.5em}
\subsection{Analysis and Discussion}
\vspace{-0.5em}
Building on the findings and insights gained through our analyses, 
we believe that the model structures of previous gait recognition methods may not be robust enough. 
This is partly due to the fact that earlier research was more lab-oriented and relied on limited datasets that did not capture the complexity of outdoor environments. 
We provide a more detailed explanation below.

\begin{figure}[tb]
\centering
\includegraphics[height=2.6cm]{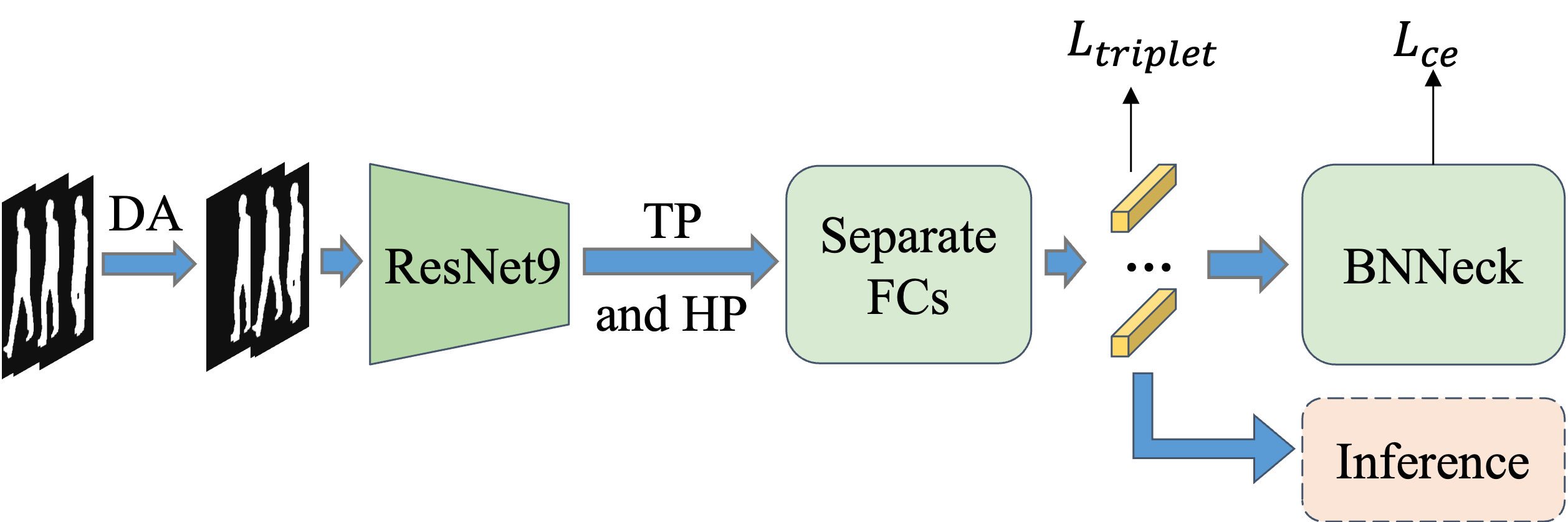}
\caption{The pipeline of GaitBase. DA is for \textit{data augmentation}, ResNet9 is for \textit{ResNet-like backbone}, TP is for \textit{temporal pooling}, HP is for \textit{horizontal pooling} (w/o pyramid structure~\cite{fu2019horizontal}).}
\label{fig:pipeline}
\vspace{-1.5em}
\end{figure}

\label{sec:weak_dataset}
\noindent \textbf{Necessity of Outdoor Evaluation.}
% As with any deep learning field, datasets are essential for deep gait recognition.
The evaluation of existing methods has primarily focused on the indoor CASIA-B~\cite{yu2006framework} and OU-MVLP~\cite{takemura2018multi}.
We argue that this approach suffers from three significant drawbacks: 
a)~\textbf{Indoor settings.} 
The walking videos are captured by an all-side camera array, and the subjects are requested to follow a particular course. 
It makes the data collection condition obviously different from the real-world scenarios.
b)~\textbf{Simple background.}
Simple laboratory background cannot reflect the complex background changes of wild scenes.
c)~\textbf {Outdated processing methods.} 
The raw RGB videos are processed by the outdated background subtraction algorithm.
% leading to many color-related background textures. 

Recently, 
new large-scale outdoor gait datasets, such as GREW~\cite{zhu2021gait} and Gait3D~\cite{zheng2022gait3d}, have emerged to promote gait recognition from in-the-lab setting to in-the-wild scenario.
However, despite these developments, most lately-published works~\cite{zhang2022adaptive,yu2022generalized,xiang2022multi} still rely excessively on indoor gait datasets.
In order to advance the development and applications of gait recognition systems, 
we suggest the community  pay more attention to outdoor datasets.

\noindent \textbf{Necessity of Comprehensive Ablation Study.}
The ablation study is recognized as the primary means of evaluating the effectiveness of individual components within a proposed method. 
However, we believe that the conclusions drawn from ablative experiments performed only on CASIA-B~\cite{yu2006framework} may have limited practical applicability. 
This is due to two primary reasons: 
a)~
CASIA-B only contains 50 subjects for evaluation. Such few testing subjects make the results vulnerable to noisy factors.
b)~
In current applications, the pedestrian segmentation is typically achieved by deep models, rather than the outdated background subtraction algorithm used by CASIA-B.

Totally, conducting a comprehensive ablation study on various large-scale datasets can provide a more robust hyper-parameter configuration and model structure.

\label{sec:base}
\noindent \textbf{Necessity of A Strong Backbone.}
The quality of a model is largely determined by the capabilities of its backbone, and a poor backbone can unjustly overestimate the effectiveness of additional modules. The evolution of CNN network models has progressed from shallow to deep, resulting in the emergence of excellent backbone networks such as AlexNet~\cite{krizhevsky2017imagenet}, VGG-16~\cite{simonyan2014very}, and ResNet~\cite{he2016deep}. However, previous works~\cite{chao2019gaitset, huang2021context,zheng2022gait3d} on gait recognition have predominantly relied on plain convolutional neural networks, consisting of several pure convolution layers. As gait recognition research advances towards applications, and larger-scale, real-world datasets~\cite{zhu2021gait,zheng2022gait3d} become more accessible, it is clear that a more robust and powerful backbone network is highly desirable to achieve accurate and reliable results.

\begin{table}[tb]
\centering
\caption{Architecture of our ResNet-like backbone.
$k, c, b$ respectively denotes kernel size, channels and number of blocks.}
\vspace{-0.5em}
\resizebox{.6\columnwidth}{!}{
\begin{tabular}{c|c}
\toprule
\multirow{2}{*}{Layer}   & Structure \\ &  [$k\times k, c$] $\times b$  \\ \hline \hline

Conv0 & [$3\times3, 64] \times 1, \textup{stride}=1$  \\ \hline
Block1      & $\begin{bmatrix}
3\times3, 64\\ 
3\times3, 64
\end{bmatrix} \times 1, \textup{stride}=1$         \\ \hline
Block2      & $\begin{bmatrix}
3\times3, 64\\ 
3\times3, 128
\end{bmatrix} \times 1, \textup{stride}=2$         \\ \hline
Block3      & $\begin{bmatrix}
3\times3, 128\\ 
3\times3, 256
\end{bmatrix} \times 1, \textup{stride}=2$         \\ \hline
Block4      & $\begin{bmatrix}
3\times3, 256\\ 
3\times3, 512
\end{bmatrix} \times 1, \textup{stride}=1$         \\ \bottomrule
\end{tabular}
}
\label{tab:resnet9}
\vspace{-1.0em}
\end{table}

\vspace{-0.5em}
\section{A Strong Baseline: GaitBase}
\vspace{-0.5em}
Using the insights gained from the above analysis, 
this section aims to offer a simple but strong baseline model to reach the totally comparable or even better performance for both indoor and outdoor evaluations. 
No bells or whistles, the obtained model should be structurally simple, experimentally powerful, and empirically robust to serve as a new baseline for further research.
To this end, we build a silhouette-based model, \textit{i.e.}, GaitBase.

\vspace{-0.5em}
\subsection{Pipeline}
\vspace{-0.5em}
We employ several easy and widely-accepted modules to form GaitBase.
As shown in Figure~\ref{fig:pipeline}, 
GaitBase follows the popular set-based and part-based paradigm and takes a ResNet-like~\cite{he2016deep} network as the backbone.
Specifically, 
the ResNet-like backbone will transform each input silhouette frame into a 3D feature map with the height, width, and channel dimensions.
And then, 
a Temporal Pooling module will aggregate the obtained sequence of feature maps by performing the maximization along the temporal dimension, 
outputting a set-level understanding of the input gait sequence, 
\textit{i.e.}, 
a 3D feature map.
Next, 
the obtained feature map will be horizontally divided into several parts, 
and each part will be pooled into a feature vector.
Accordingly, 
we get several feature vectors and further use a separate fully connected layer to map them into the metric space.
Finally, 
we employ the widely-used BNNeck~\cite{luo2019bag} to adjust feature space, 
and the separate triplet and cross-entropy losses  are utilized to drive the training process. The overall loss function
is formulated by $L = L_{triplet}+L_{ce}$.

\vspace{-0.5em}
\subsection{ResNet-like Backbone}
\vspace{-0.5em}
ResNet~\cite{he2016deep} is one of the most successful deep models and has been broadly used as the backbone for many vision tasks.
As for gait recognition, 
most of the current works still take a shallow stack of several standard or customized convolution layers as the backbone.
In this paper, 
we develop a ResNet-like network to serve as the backbone of GaitBase.
As shown in Table~\ref{tab:resnet9}, 
the network comprises initial convolution layer followed by a stack of four basic residual blocks.
All the layers are equipped with BN~\cite{ioffe2015batch} and ReLU activation layers which are skipped for conciseness.

\begin{table}[tb]
\centering
\caption{The amount of the identities (\#ID) and sequences (\#Seq) covered by the employed datasets.}
\vspace{-0.5em}
\resizebox{1.0\columnwidth}{!}{
\begin{tabular}{cccccc}
% \hline
\toprule
\multirow{2}{*}{Dataset} & \multicolumn{2}{c}{Train Set} & \multicolumn{2}{c}{Test Set} & \multirow{2}{*}{Condition} \\
                         & \#Id            & \#Seq           & \#Id           & \#Seq           &                         \\ \midrule %\hline
CASIA-B                  & 74            & 8,140         & 50           & 5,500         & NM, BG, CL              \\
OU-MVLP                   & 5,153         & 144,284       & 5,154        & 144,312       & NM                      \\
GREW                     & 20,000        & 102,887       & 6,000        & 24,000        & Diverse                   \\
Gait3D                   & 3,000         & 18,940        & 1,000        & 6,369         & Diverse                   \\ \bottomrule %\hline
\end{tabular}
}
\label{tab:datasets}
\vspace{-1.0em}
\end{table}

\vspace{-0.5em}
\subsection{Data Augmentation}
\vspace{-0.5em}
In order to match the practical usage, we explore the data augmentation (DA) strategy  specific to gait silhouettes that previous works have seldom considered.
Our experiments show that several spatial augmentation operations, such as horizontal flipping, rotation, perspective, and affine transformations, can  improve recognition accuracy.
Additionally, 
we discover that allowing for an unfixed length of the input gait sequence during the training phase can further enhance the final performance.

\vspace{-0.5em}
\subsection{Comparison with Other Methods}
\vspace{-0.5em}
The driving philosophy behind the development of GaitBase is the belief that simplicity is beautiful.
Accordingly, we adopt several succinct, widespread, and validated techniques to make GaitBase robust and efficient. 
These include treating a gait sequence as a set, using part-based models for human body description, utilizing the widely used ResNet-like network as a backbone, and leveraging the general data augmentation strategy to fit the practical usage.
These settings do not require any unique designs but still reach comparable or even better performance on the existing gait datasets, regardless of indoor or outdoor ones.

\vspace{-0.5em}
\section{Experiment}
\vspace{-0.5em}
\label{sec:exp}
In this section, we mainly present a comprehensive study to show the effectiveness and robustness of GaitBase as well as its components on various gait datasets.

\vspace{-0.5em}
\subsection{Datasets}
\vspace{-0.5em}
Four public gait datasets are utilized, involving the most widely-used indoor datasets, CASIA-B~\cite{yu2006framework} and OU-MVLP~\cite{takemura2018multi}, and the largest gait dataset in the wild, GREW~\cite{zhu2021gait} and Gait3D~\cite{zheng2022gait3d}.
Table~\ref{tab:datasets} displays the statistics on the number of identities and sequences that are covered in each of these datasets.
The following section provides a detailed overview of the collection process for each of these datasets, emphasizing the significant differences that exist between indoor and outdoor datasets.

% \noindent
\textbf{CASIA-B.} 
Each subject is asked to follow a particular course with three walking conditions, \textit{i.e.}, normal walking (NM), walking with bags (BG), and walking with coats (CL). The obtained videos were captured by an all-side camera array with 11 filming viewpoints. 
The gait silhouettes were generated through the old background subtraction algorithm.

\begin{table}[tb]
\centering
\caption{Implementation details.
The batch size $(q, k)$ indicates $q$ subjects with $k$ sequences per subject.}
\vspace{-0.5em}
\resizebox{0.85\columnwidth}{!}{
\begin{tabular}{cccc}
% \hline
\toprule
 \multirow{2}{*}{DataSet} & \multirow{2}{*}{Batch Size} & \multirow{2}{*}{MultiStep Scheduler} & \multirow{2}{*}{Steps} \\
                          &                             &                             &                        \\ \midrule %\hline
 CASIA-B                  & (8, 16)                    & (20k, 40k, 50k)            & 60k                    \\
 OU-MVLP                  & (32, 8)                    & (60k, 80k, 100k)             & 120k                    \\
 GREW                     & (32, 4)                    & (80k, 120k, 150k)             & 180k                    \\
 Gait3D                   & (32, 4)                    & (20k, 40K, 50k)               & 60k                    \\ \bottomrule %\hline                 \\ \ %\hline
\end{tabular}
}
\label{tab:implementation}
\vspace{-1.0em}
\end{table}

\begin{table*}[tb]
\centering
\caption{
The performance comparison on four major datasets: rank-1 accuracy (\%) are reported. CASIA-B*~\protect\cite{liang2022gaitedge} is a re-processed version of CASIA-B. 
The highest two recognition results are in \textbf{bold}.
DA is for \textit{data augmentation}.
\vspace{-0.5em}
}
\resizebox{1.8\columnwidth}{!}{%
\begin{tabular}{c|cc|ccccccccc}
% \hline
\toprule
\multirow{3}{*}{Method}          & \multicolumn{2}{c|}{\multirow{2}{*}{\begin{tabular}[c]{@{}c@{}}Condition\end{tabular}}} & \multicolumn{9}{c}{Dataset}                                                                                                                                                                                                                  \\ \cline{4-12} 
                                 & \multicolumn{2}{c|}{}                                                                                    & \multicolumn{3}{c|}{CASIA-B}                              & \multicolumn{3}{c|}{CASIA-B*}                             & \multicolumn{1}{c|}{\multirow{2}{*}{OU-MVLP}} & \multicolumn{1}{c|}{\multirow{2}{*}{GREW}} & \multirow{2}{*}{Gait3D} \\ \cline{2-9}
                                 & \multicolumn{1}{c|}{BNNeck~\cite{luo2019bag}}                                 & DA                                         & NM            & BG            & \multicolumn{1}{c|}{CL}   & NM            & BG            & \multicolumn{1}{c|}{CL}   & \multicolumn{1}{c|}{}                         & \multicolumn{1}{c|}{}                      &                         \\ \hline \hline
\multirow{3}{*}{GaitBase (Ours)} & $\times$                                                    & $\times$                                   & 96.7          & 91.9          & \multicolumn{1}{c|}{74.9} & 94.3          & 87.9          & \multicolumn{1}{c|}{73.5} & \multicolumn{1}{c|}{89.6}                     & \multicolumn{1}{c|}{54.7}                  & 56.9                    \\
 &             $\times$                                      &       \checkmark                             & 97.8         &93.9           & \multicolumn{1}{c|}{77.6} & \textbf{96.2}          & \textbf{91.4}          & \multicolumn{1}{c|}{77.3} & \multicolumn{1}{c|}{88.3}                     & \multicolumn{1}{c|}{\textbf{57.9}}                  & \textbf{62.0}                \\
                                 & \checkmark                                                  & $\times$                                   & 97.3          & 92.9          & \multicolumn{1}{c|}{78.0} & 94.5          & 90.0          & \multicolumn{1}{c|}{75.9} & \multicolumn{1}{c|}{\textbf{90.8}}                     & \multicolumn{1}{c|}{57.7}                  & 54.7                    \\
                                 & \checkmark                                                  & \checkmark                                 & 97.6          & 94.0          & \multicolumn{1}{c|}{77.4} & \textbf{96.5} & \textbf{91.5} & \multicolumn{1}{c|}{\textbf{78.0}} & \multicolumn{1}{c|}{90.0}                     & \multicolumn{1}{c|}{\textbf{60.1}}                  & \textbf{64.6}           \\ \hline \hline
GaitSet~\cite{chao2019gaitset}                    & \multicolumn{2}{c|}{AAAI 2019}                                                                           & 95.8          & 90.0          & \multicolumn{1}{c|}{75.4} & 92.3          & 86.1          & \multicolumn{1}{c|}{73.4} & \multicolumn{1}{c|}{87.1}                     & \multicolumn{1}{c|}{48.4}                  & 36.7                    \\
GaitPart~\cite{fan2020gaitpart}                   & \multicolumn{2}{c|}{CVPR 2020}                                                                           & 96.1          & 90.7          & \multicolumn{1}{c|}{78.7} & 93.1          & 86.0          & \multicolumn{1}{c|}{75.1} & \multicolumn{1}{c|}{88.7}                     & \multicolumn{1}{c|}{47.6}                  & 28.2                    \\
GaitGL~\cite{lin2021gait}                     & \multicolumn{2}{c|}{ICCV 2021}                                                                           & 97.4          & 94.5          & \multicolumn{1}{c|}{83.8} & 94.1          & 90.0          & \multicolumn{1}{c|}{\textbf{81.4}} & \multicolumn{1}{c|}{89.7}                     & \multicolumn{1}{c|}{47.3}                  & 29.7                    \\
CSTL~\cite{huang2021context}                       & \multicolumn{2}{c|}{ICCV 2021}                                                                           & \textbf{98.0}          & \textbf{95.4}          & \multicolumn{1}{c|}{\textbf{87.0}} & \multicolumn{3}{c|}{-}                                    & \multicolumn{1}{c|}{90.2}                     & \multicolumn{1}{c|}{50.6}                  & 11.7                    \\
3DLocal~\cite{huang20213d}                    & \multicolumn{2}{c|}{ICCV 2021}                                                                           & \textbf{98.3} & \textbf{95.5} & \multicolumn{1}{c|}{\textbf{84.5}} & \multicolumn{3}{c|}{-}                                    & \multicolumn{1}{c|}{\textbf{90.9}}                     & \multicolumn{1}{c|}{-}                     & -                       \\
SMPLGait~\cite{zheng2022gait3d}                   & \multicolumn{2}{c|}{CVPR 2022}                                                                           & \multicolumn{3}{c|}{-}                                    & \multicolumn{3}{c|}{-}                                    & \multicolumn{1}{c|}{-}                        & \multicolumn{1}{c|}{-}                     & 46.3                    \\ \bottomrule %\hline
\end{tabular}
}
\label{tab:overview}
\vspace{-1.0em}
\end{table*}

% \noindent
\textbf{OU-MVLP}
is one of the largest public gait datasets.
However, similar to CASIA-B, each subject in OU-MVLP was required to walk in a fixed course with fixed cameras and produce the gait silhouette by the outdated background subtraction algorithm.
OU-MVLP contains only one walking status for each subject and thus lacks the clothing and carrying changes, 
making the recognition task relatively easy.

% \noindent
\textbf{GREW}
is the largest gait dataset in the wild up to date, to our knowledge. 
Its raw videos are collected from 882 cameras in a large public area, containing nearly 3,500 hours of 1,080$\times$1,920 streams.
Besides the tens of thousands of identities, many other human attributes have been annotated, 
\textit{e.g.}, 2 gender, 14 age groups, 5 carrying conditions, and 6 dressing styles.
Therefore, GREW is believed to include adequate and diverse practical variations.

% \noindent
\textbf{Gait3D} is a large-scale  gait dataset in the wild.
It
was collected in a supermarket and contains 1,090 hours of videos with 1,920$\times$1,080 resolution and 25 FPS.

Our implementation follows official protocols, including the training/testing and gallery/probe set partition strategies. 
Rank-1 accuracy is used as the primary evaluation metric.

\vspace{-0.5em}
\subsection{Implementation Details}
\vspace{-0.5em}
Table~\ref{tab:implementation} displays the main hyper-parameters of our experiments.
All the source codes are based on OpenGait, which is proposed in this work. 
The input resolution for all datasets is set to $64 \times 44$. We set the triplet loss margin as 0.2, the initial learning rate and the weight decay of SGD optimizer~\cite{robbins1951stochastic} as 0.1 and  0.0005, respectively. 
The input frame number is unfixed.
Furthermore, we employ Horizontal Flipping, Random Erasing~\cite{zhong2020random}, Rotation, Perspective Transformation, and Affine Transformation as spatial data augmentation techniques.
Their details are presented in the supplementary.
% We present more their details in the supplementary material.

\vspace{-0.5em}
\subsection{Comparison with Other State-of-the-Arts}
\vspace{-0.5em}
We compare our GaitBase with other published state-of-the-art methods, 
\textit{e.g.}, 
GaitSet~\cite{chao2019gaitset}, GaitPart~\cite{fan2020gaitpart}, GaitGL~\cite{lin2021gait}, CSTL~\cite{huang2021context}, 3DLocal~\cite{huang20213d}, and SMPLGait~\cite{zheng2022gait3d}, 
on the various popular datasets, 
\textit{i.e.}, 
CASIA-B~\cite{yu2006framework}, CASIA-B*~\cite{liang2022gaitedge}, OU-MVLP~\cite{takemura2018multi}, GREW~\cite{zhu2021gait} and Gait3D~\cite{zheng2022gait3d}.
Among these datasets, 
CASIA-B*~\cite{liang2022gaitedge} is a re-segmented version of CASIA-B.

As shown in Table~\ref{tab:overview}, 
excluding CASIA-B, 
GaitBase achieves a totally competitive or even significantly better performance than other state-of-the-art methods.
More importantly, 
GaitBase exceeds other methods by a more considerable margin on the datasets captured from practical scenarios, 
\textit{i.e.}, 
+9.5\% on GREW and +18.3\% on Gait3D, 
richly exhibiting both the effectiveness and robustness of our strong baseline at the time of dealing the troublesome real-world noisy factors.

As for CASIA-B, our GaitBase achieves a less competitive accuracy. It illustrates that previous methods are finely designed for the CASIA-B dataset to achieve high accuracy while ignoring its bias from wild data.
% The main reason is that CASIA-B is far from the current real scene dataset and is no longer a reasonable dataset for evaluation. 
Additionally, our method still achieves relatively high performance on re-segmented CASIA-B*.

\vspace{-0.5em}
\subsection{Ablation Study}
\vspace{-0.5em}
To study the effectiveness of BNNeck and DA, we conduct additional experiments on all the mentioned datasets, as shown in Table \ref{tab:overview}. 

\noindent
\textbf{Impact of BNNeck.}
It should be noted that GaitBase with BNNeck outperforms GaitBase without BNNeck on almost all datasets, which shows that BNNeck generally plays a positive role in GaitBase.

\noindent
\textbf{Impact of DA.}
The results show that with data augmentation, the performance of GaitBase  on the outdoor datasets is improved (+2.4\% for GREW and +9.9\% for Gait3D). 
In comparison, the performance on the indoor datasets is degraded (-0.8\% for OU-MVLP). 
This observation indicates that our data augmentation strategy is more capable of suppressing over-fitting caused by the noisy variations in practical scenarios.

\noindent
\textbf{Impact of ResNet9.} 
We have also tried the standard ResNet50~\cite{he2016deep} as the backbone.
Further ablative experiments show that ResNet50 does not bring significant performance improvement on Gait3D (+1.9\%) but causes serious over-fitting on CASIA-B* (converges better but performs worse, \textit{i.e.}, -5.2\%, -6.8\%, -9.6\% on NM, BG and CL). Moreover, using ResNet50 also results in over $3\times$ computation cost (3.939 \textit{v.s.} 1.183 GFLOPs per silhouette image). Since the primary goal of GaitBase is to provide a practical yet powerful baseline, we finally choose the simple ResNet9. 
% Admittedly, exploring a better deep model for gait recognition has a long way to go, and we will offer more discussions in the supplementary material.

\vspace{-1.0em}
\section{Conclusion}
\vspace{-0.5em}
This paper presents a codebase, OpenGait, developed for deep gait recognition. OpenGait provides a fair and easy-to-use platform for future gait recognition works, facilitating researchers to implement new ideas more efficiently. Subsequently, we implemented most state-of-the-art methods in OpenGait and compared them fairly. Some insights different from those in the original papers have been found in our fair ablation studies. Based on those findings and understandings, a simple but efficient gait recognition model, GaitBase, is proposed. GaitBase can achieve the best performance on most gait datasets, especially on datasets in the wild. So GaitBase can serve as a new baseline for future studies. 
% We sincerely hope OpenGait and GaitBase can help researchers in this field by providing a fair codebase and a good baseline method.

% \noindent \textbf{Ethical Statements.}
% We are highly concerned about personal information security and argue that the improper use or abuse of gait recognition will threaten personal privacy.
% We believe that the development of vision techniques should only devote to the cause of human happiness.

\noindent \textbf{Acknowledgement.}
This work was supported in part by the National Natural Science Foundation of China under Grant 61976144, Grant 62276025, and Grant 62206022, 
in part by the Stable Support Plan Program of Shenzhen Natural Science Fund under Grant 20200925155017002, 
in part by the National Key Research and Development Program of China under Grant 2020AAA0140002, 
and in part by the Shenzhen Technology Plan Program under Grant JSGG20201103091002008, and Grant KQTD20170331093217368.

\newpage
%%%%%%%%% REFERENCES
{\small
\bibliographystyle{ieee_fullname}
\bibliography{egbib}
}

\newpage
\section{Supplementary Material}
The source code of GaitBase is avaliable at \url{https://github.com/ShiqiYu/OpenGait}.
In this section, we first explore the effectiveness of several usual spatial data augmentation operations. 
Then, we conduct comprehensive experiments to analyze the effect of random training input length. Lastly, we talk about some future works that are worth further exploration. 

\subsection{Effect of Spatial Data Augmentation}
\begin{figure}[htbp]
\centering
\includegraphics[height=5.0cm]{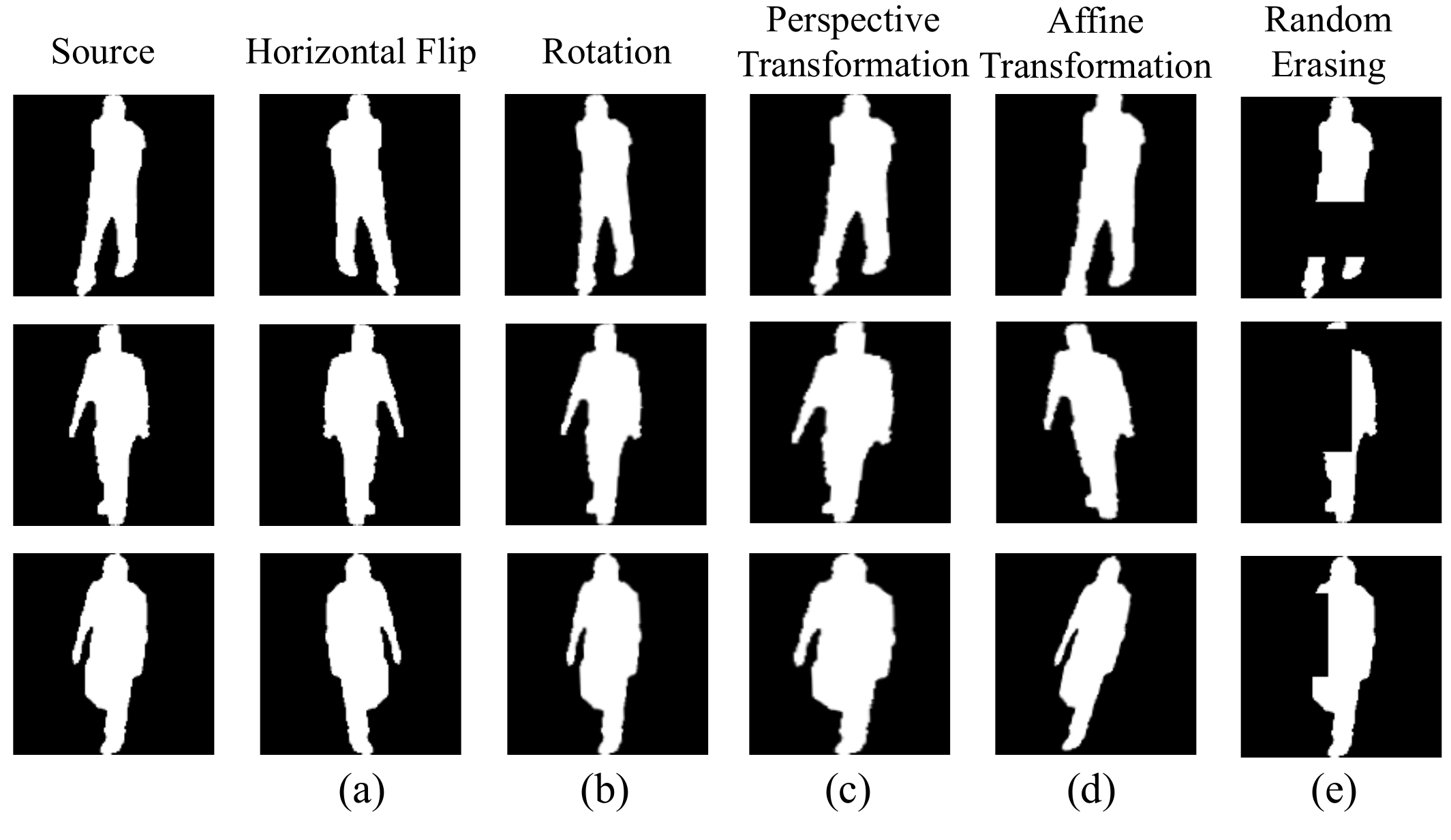}
\caption{The visualization of source image with different spatial data augmentation operations. 
In Rotation, the twist angle is randomly sampled from  $[-10^{\circ}, +10^{\circ}]$. 
In Perspective Transformation, 
the source axes are randomly skewed within 10 pixels to produce the transform axes of perspective. 
In Affine Transformation, we perform Rotation on source images and then shear them with their level ranging on $[-5\times10^{-3}, 5\times10^{-3}]$. 
In Random Erasing, we use the original hyper-parameters~\cite{zhong2017re} for image classification. 
For each input sequence, the probability of performing spatial augmentation operations is set to 0.5.
% ~\cite{zhong2020random}
}
\label{fig:da-vis}
\end{figure}

% In this work, we consider several data augmentation methods to enhance the proposed baseline model, GaitBase. 
As shown in  Fig.~\ref{fig:da-vis}, 
we perform various spatial augmentation techniques to enlarge the data space and avoid overfitting of the model.
We conduct an ablation study on two commonly used indoor and outdoor datasets, \textit{i.e.}, CASIA-B*\footnote{The conclusions obtained from the experiments on CASIA-B and CASIA-B* are consistent. Here we only present the results on CASIA-B* for brevity.} and Gait3D, to evaluate the efficacy of these approaches experimentally. 
The results are shown in Table \ref{tab:da-result}. 

\begin{table}[t]
\centering
\caption{Ablation study for spatial data augmentation with the fixed length training input. 
Rank-1 accuracies (\%) are reported on CASIA-B* and Gait3D, HF for Horizontal Flip, R for Rotation, PT for Perspective Transformation, AT for Affine Transformation, and RE for Random Erasing. 
The bracket indicates that the performance outperforms the GaitBase without data augmentation. }
\resizebox{\columnwidth}{!}{%
\begin{tabular}{c|ccccc|c|c} 
\hline
\multirow{2}{*}{Group} & \multicolumn{5}{c|}{Data Augmentation} & \multirow{2}{*}{CASIA-B*} & \multirow{2}{*}{Gait3D}  \\ 
\cline{2-6}
                        & HF & R & PT & AT & RE                  &                           &                          \\ 
\hline
-                &    &   &    &    &                     & 86.8                      & 54.7                     \\ 
\hline\hline
(a)                     &  \checkmark  &   &    &    &                     & 86.5                      & (59.5)                   \\ 
\hline
(b)                     &    &  \checkmark &    &    &                     & (87.4)                    & (60.5)                   \\ 
\hline
(c)                     &    &   & \checkmark   &    &                     & 86.6                      & (58.9)                   \\ 
\hline
(d)                     &    &   &    &  \checkmark  &                     & 79.0                      & (55.8)                     \\ 
\hline
(e)                     &    &   &    &    &    \checkmark                 & (87.8)                    & 54.1                     \\ 
\hline
(f)                     &    &  \checkmark &    &    &               \checkmark      & \textbf{88.7}             & -                        \\ 
\hline
(g)                     &   \checkmark &  \checkmark &  \checkmark  &    &                     & -                         & \textbf{(62.4)}          \\
\hline
\end{tabular}
}
\label{tab:da-result}
\end{table}

Horizontal Flip can largely simulate a mirror transformation of the filming viewpoint.
In (a), we observe that although it fails to improve the performance of CASIA-B*, it significantly improves accuracy on Gait3D. 
This can be explained by the fact that CASIA-B* is recorded in a laboratory environment using an all-sided camera array, while Gait3D is captured in real-world conditions with comparatively fewer viewpoint changes per subject.

From the experiment (b) in Table \ref{tab:da-result}, Rotation technique slightly benefits both CASIA-B* and Gait3D. 

Perspective Transformation aims to simulate the effects of different camera heights. 
As shown in (c), our experiments indicate that this technique only has a significant impact on Gait3D dataset. 
The cause should be that there are no camera height changes in CASIA-B*, whereas there are such changes in Gait3D.

From the experiment (d) in Table \ref{tab:da-result}, it appears that Affine Transformation is not able to effectively simulate the noisy factors present in both indoor CASIA-B* and outdoor Gait3D datasets, thereby failing to bring any performance gain on these datasets.
% has no gain on both datasets.  CASIA-B(*) is too clean and easy, while Gait3D has much occlusion. 

The main goal of Random Erasing~\cite{zhong2017re} is to simulate the occlusion conditions and avoid the over-fitting problem in the spatial dimension.
From (e), we note that the erasing operation is challenging to simulate the practical occlusion cases and thus makes almost no difference on Gait3D. 

Building upon the above experimental results, we apply the Rotation and Random Erasing for the indoor datasets such as CASIA-B*, CASIA-B and OU-MVLP.
On the other hand, for outdoor datasets like Gait3D and GREW, 
we employ the combination of Horizontal Flip, Rotation, and Perspective Transformation as the augmentation strategy.
As evident from (f) and (g), it can be observed that the data augmentation approach brings accuracy improvements of 1.9\% and 7.7\% on CASIA-B* and Gait3D, respectively.

Based on the above analysis, we again expose the sizable gap between the experimental and practical gait data.
Therefore, we propose that the research community should focus more attention on outdoor gait datasets for better practicality. 

\begin{figure}[t]
\centering
\includegraphics[height=6.3cm]{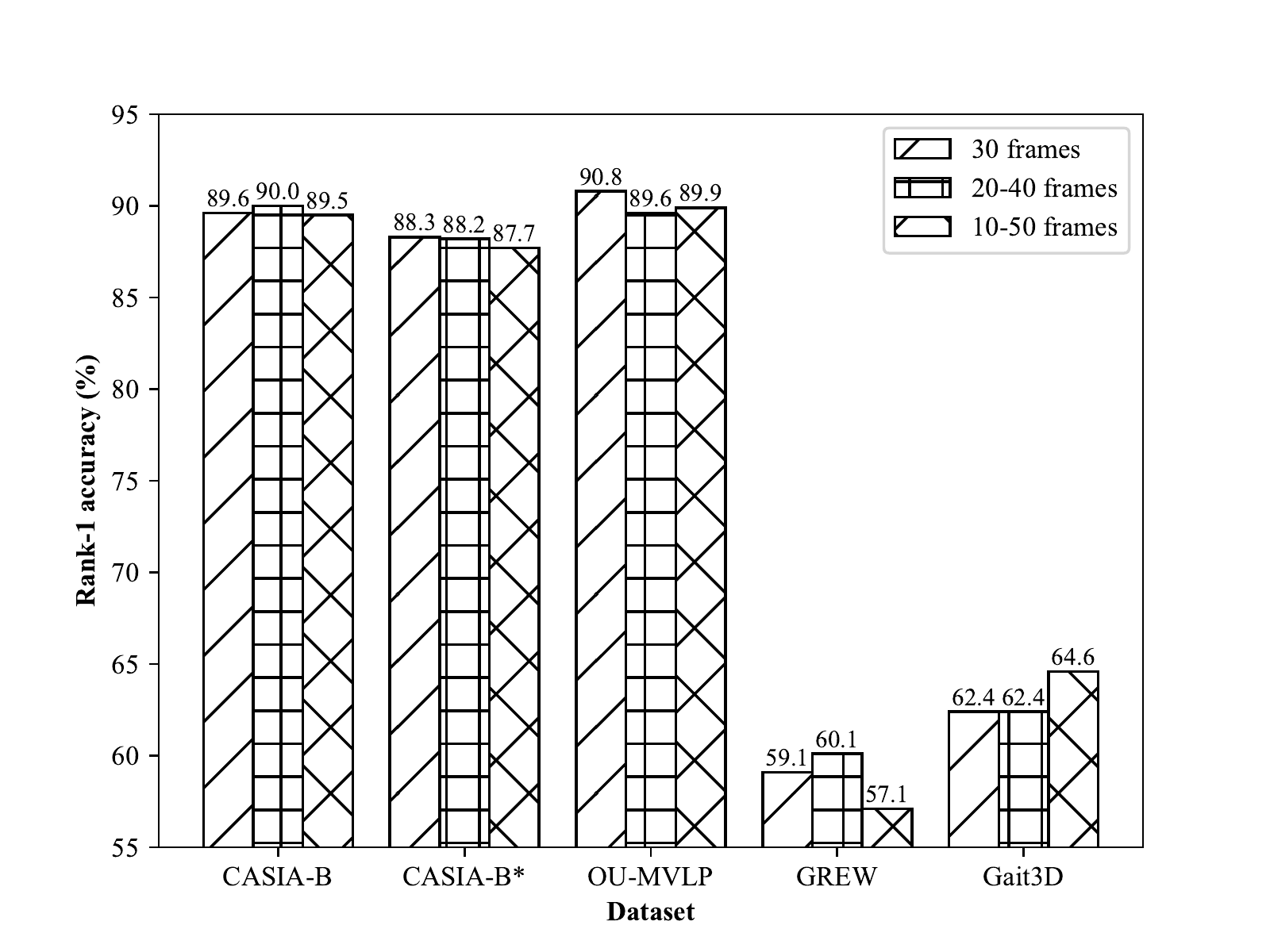}
\caption{The effect of random  training input length. 
20-40 and 10-50 represent that the input lengths are uniformly distributed over 20 to 40 and 10 to 50.}
\label{fig:da-frame}
\end{figure}

\subsection{Effect of Random Training Input Length}
In this subsection, we investigate the effect of random training input length on the final recognition performance.
As shown in Fig.~\ref{fig:da-frame}, we observe that the fixed length input works relatively optimal for the indoor dataset such as CASIA-B, CASIA-B* and OU-MVLP.
On the other hand, the use of random length input yields superior performance for outdoor datasets, such as GREW and Gait3D. 
Theoretically, similar to the popular Dropout~\cite{srivastava2014dropout} technique, the usage of random sequence length can maintain consistency in features, regardless of the input length, thus easing the over-fitting problem in the temporal dimension. 
In laboratory-acquired datasets, frame dropping and walking speed fluctuations are minimal, thereby resulting in a uniform number of frames in the gait cycle.
As a result, the random training input length has little impact on indoor datasets such as CASIA-B, CASIA-B*, and OU-MVLP, which may be attributed to this consistent nature.

\subsection{Future Work and Discussion}
This paper presents a comprehensive benchmark study towards gait recognition applications, which includes a flexible codebase, a series of experimental reviews, and a robust baseline.
In addition, here we highlight some subsequent works that are worth further exploration. 

\noindent \textbf{Gait Verification Task.}
The evaluation protocols of existing gait datasets mostly focus on identification (retrieval) tasks, resulting in gait verification scenarios being ignored in most cases.
Typically, there are two categories of methods for performing verification processes: training a binary classifier~\cite{wu2016comprehensive, zhang2019learning} or inferring a conclusive distance threshold to determine whether the two subjects come from the same identity. 
However, since clothing and viewpoint changes may dramatically impact gait appearance, reducing the intra-class distance is always a challenging issue for gait recognition. 
This poses a huge obstacle for gait verification applications. 
We encourage the research community to devote more attention to this complex topic, as it is widely needed for practical usage.

\noindent \textbf{Stronger Baseline Model.} 
Although the proposed baseline model, GaitBase, has achieved state-of-the-art performance on the largest outdoor gait dataset, GREW~\cite{zhu2021gait}, with a rank-1 accuracy of 60.1\%, there is still a significant gap in achieving an accurate enough gait recognition for real-world applications. 
Additionally, there has been a modeling shift from CNNs to Transformers~\cite{vaswani2017attention, dosovitskiy2020image, liu2021swin} in many visual tasks. 
With its outstanding modeling capabilities, transformer-based gait recognition offers a fascinating solution to the challenges posed by outdoor environments, yet it has not received the attention it deserves.
Therefore, the development of a stronger baseline model, such as a transformer-based model, remains a pressing issue for practical gait recognition.

\noindent \textbf{Unsupervised Gait Recognition.}
The large-scale collection of annotated gait data in the wild is economically expensive and usually limited in the trade-off between the diversity and scale.
For example, the largest outdoor gait dataset, GREW~\cite{zhu2021gait}, covers over 20,000 subjects,
but each subject, on average, only has about 4.5 walking sequences mostly captured from nearly front and back viewpoints.
Additionally, it is challenging for outdoor gait datasets like GREW to include the long-term changes in clothing, age, hair, and body sizes for each subject as their collection process typically finishes in several months. 
Therefore, we consider learning the general gait representation, \textit{i.e.}, prior identity knowledge, from unlabelled walking videos to be a challenging yet highly appealing task for further study.

\section*{Ethical Statements.}
We are highly concerned about personal information security and argue that the improper use or abuse of gait recognition will threaten personal privacy.
We believe that the development of vision techniques should only devote to the cause of human happiness.

\section*{Acknowledgment}
We want to thank the reviewers for their efforts and the authors of the references for their inspiring insights and awesome achievements.

% \newpage
% %%%%%% REFERENCES
% {\small
% \bibliographystyle{ieee_fullname}
% \bibliography{egbib}
% }

\end{document}